\documentclass[runningheads]{llncs}
\usepackage[T1]{fontenc}

\usepackage{graphicx,verbatim}
\usepackage{amsmath} 
\usepackage{graphicx}
\usepackage{amssymb}
\usepackage{booktabs}
\usepackage{multirow}
\usepackage{graphicx}
\usepackage{xcolor}
\usepackage{colortbl} % 导入 colortbl 包
\usepackage{tabularx}
\usepackage{xtab}
\usepackage{verbatim} 
\usepackage{enumitem}
\usepackage{makecell}
\usepackage{pifont} 
\usepackage{ulem}
\usepackage{hyperref}
\begin{document}
\title{ WSI-INR: Implicit Neural Representations for Lesion Segmentation in Whole-Slide Images}
\titlerunning{Implicit Neural Representations for Lesion Segmentation in WSI}
\begin{comment}  %% Removed for anonymized MICCAI submission
\author{First Author\inst{1}\orcidID{0000-1111-2222-3333} \and
Second Author\inst{2,3}\orcidID{1111-2222-3333-4444} \and
Third Author\inst{3}\orcidID{2222--3333-4444-5555}}
%
\authorrunning{F. Author et al.}
\institute{Princeton University, Princeton NJ 08544, USA \and
Springer Heidelberg, Tiergartenstr. 17, 69121 Heidelberg, Germany
\email{lncs@springer.com}\\
\url{http://www.springer.com/gp/computer-science/lncs} \and
ABC Institute, Rupert-Karls-University Heidelberg, Heidelberg, Germany\\
\email{\{abc,lncs\}@uni-heidelberg.de}}

\end{comment}

% \author{Anonymized Authors}  %% Added for anonymized MICCAI submission
% \authorrunning{Anonymized Author et al.}
% \institute{Anonymized Affiliations \\
%     \email{email@anonymized.com}}

\author{Yunheng Wu\inst{1}\and
Wenqi Huang\inst{2,3} \and
Liangyi Wang\inst{1}\and \\
Masahiro Oda\inst{4,1}\and 
Yuichiro Hayashi\inst{1}\and \\
Daniel Rueckert\inst{2,3,5,6}\and
Kensaku Mori\inst{1,4,7}}
\authorrunning{Yunheng Wu et al.}

\institute{Graduate School of Informatics, Nagoya University, Nagoya, Japan
\email{yunhengwu@mori.m.is.nagoya-u.ac.jp; kensaku@is.nagoya-u.ac.jp}\\
\and Chair for AI in Healthcare and Medicine, Technical University of Munich (TUM) and TUM University Hospital, Munich, Germany
\and School of Computation and Information Technology,\\ Technical University of Munich, Munich, Germany
\and Information Technology Center, Nagoya University, Nagoya, Japan
\and Munich Center for Machine Learning,\\ Technical University of Munich, Munich, Germany
\and Department of Computing, Imperial College London, London, United Kingdom
\and Research Center for Medical Bigdata,\\ National Institute of Informatics, Tokyo, Japan
}

\maketitle              % typeset the header of the contribution
\begin{abstract}
Whole-slide images (WSIs) are fundamental for computational pathology, where accurate lesion segmentation is critical for clinical decision making. Existing methods partition WSIs into discrete patches, disrupting spatial continuity and treating multi-resolution views as independent samples, which leads to spatially fragmented segmentation and reduced robustness to resolution variations. To address the issues, we propose WSI-INR, a novel patch-free framework based on Implicit Neural Representations (INRs). WSI-INR models the WSI as a continuous implicit function mapping spatial coordinates directly to tissue semantics features, outputting segmentation results while preserving intrinsic spatial information across the entire slide. In the WSI-INR, we incorporate multi-resolution hash grid encoding to regard different resolution levels as varying sampling densities of the same continuous tissue, achieving a consistent feature representation across resolutions. In addition, by jointly training a shared INR decoder, WSI-INR can capture general priors across different cases. Experimental results showed that WSI-INR maintains robust segmentation performance across resolutions; at Base/4, our resolution-specific optimization improves Dice score by +26.11\%, while U-Net and TransUNet decrease by 54.28\% and 36.18\%, respectively. Crucially, this work enables INRs to segment highly heterogeneous pathological lesions beyond structurally consistent anatomical tissues, offering a fresh perspective for pathological analysis. 

% Authors must provide keywords and are not allowed to remove this Keyword section.
% \and Digital Pathology\and Lesion Segmentation 

\end{abstract}
\section{Introduction}

\begin{figure}[tb]
  \centering
  \includegraphics[page=1, width=\textwidth, clip, trim=0cm 0.6cm 0cm 2.3cm ]{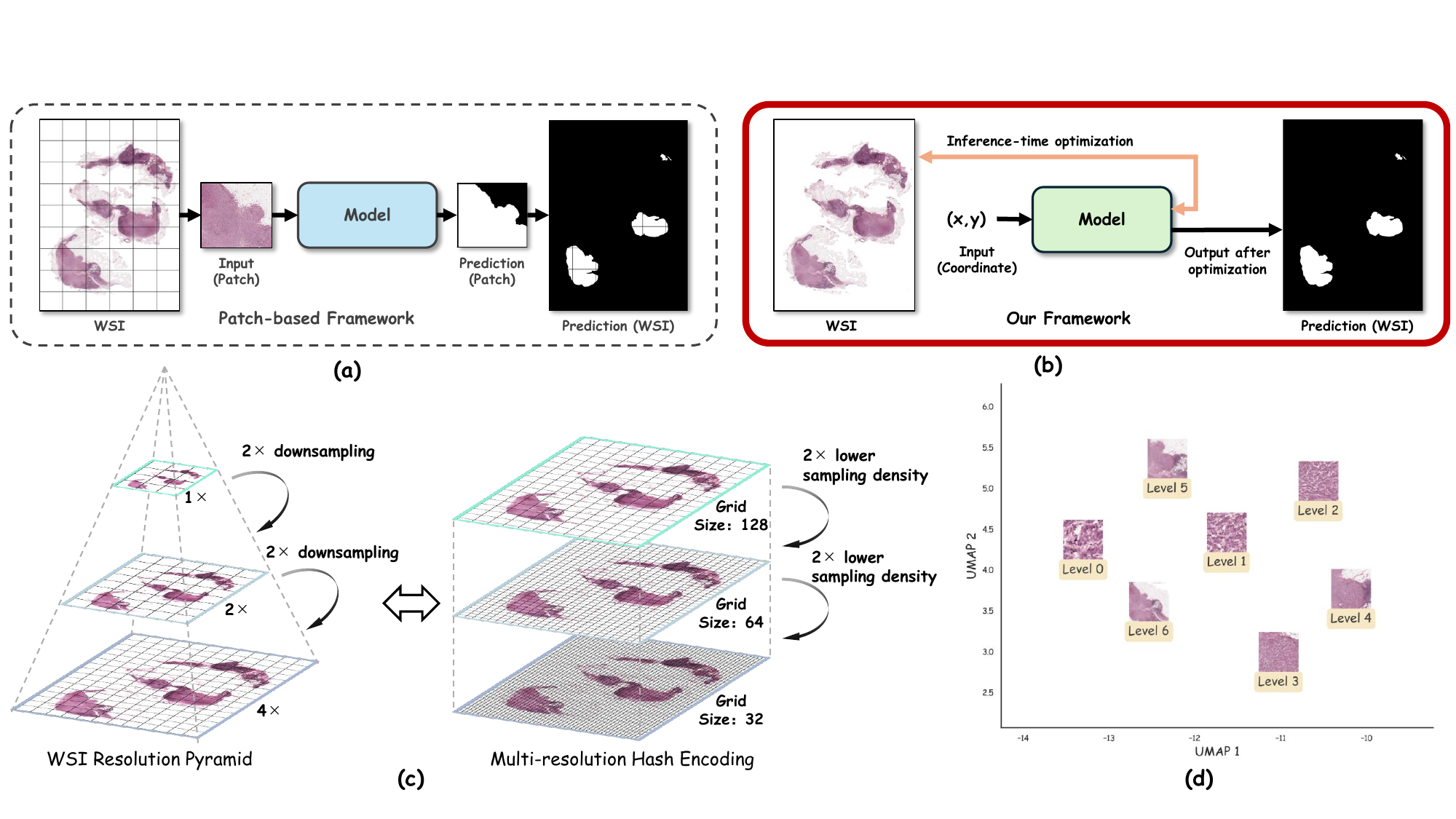}
  \caption{(a) Patch-based methods. (b) Our WSI-INR. (c) WSI resolution pyramid and multi-resolution hash encoding. (d) Patches across resolutions exhibit inconsistent representations. Visualization of U-Net bottleneck features trained on a WSI dataset.}
  \label{figure1}
\end{figure}

Whole-slide images (WSIs) are a core data modality in digital pathology, acquired by scanning entire tissue slides at high magnification and stored in a multi-resolution pyramid structure~\cite{farahani2015whole,srinidhi2021deep,campanella2019clinical,shao2021transmil,li2022comprehensive,brixtel2022whole}. A large-scale high-resolution WSI preserves tissue architecture and cellular morphology across multiple spatial scales, enabling lesion segmentation to automatically localize lesions and assist diagnosis~\cite{li2022comprehensive,brixtel2022whole}. However, the level of resolution makes direct computation on a whole slide impractical. Therefore, most existing methods tile a WSI into a large set of patches (Fig.~\ref{figure1}-(a)), converting whole-slide modeling to the processing of discrete patches. This leads to \textbf{two key issues:} (1) WSIs are divided into patches, breaking the spatial continuity of tissue structures. Each patch is encoded independently, and the global structure is typically approximated only through subsequent feature aggregation~\cite{campanella2019clinical,lu2021data,xu2019camel} or attention mechanisms~\cite{tokunaga2019adaptive,campanella2019clinical,shao2021transmil}. However, such a discretized representation establishes correlations merely at the feature level and does not directly model global spatial information, making it difficult for the model to capture the true spatial information of lesions. (2) In pathological diagnosis, pathologists examine the same spatial location at multiple resolutions. In addition, WSIs acquired at different institutions may have different sampling densities due to variations in scanners. In practical applications, storage capacity constraints may require models to be trained at one resolution but deployed at another. Existing patch-based methods treat WSIs at different resolutions as independent samples and rely on multi-scale feature fusion~\cite{das2017classifying,ronneberger2015u}. However, different WSI resolutions represent multi-scale observations of the same tissue rather than different tissues. Patches at the same spatial coordinate share identical pixel size but differ in physical field of view and visual features. Independent training makes the model treat sampling-scale variations as semantic differences, leading to fragmented structures during cross-resolution inference (Fig.~\ref{figure1}-(d)). Moreover, using patches from multiple resolutions significantly increases computational cost and training time.

As a different modeling framework, implicit neural representations (INRs) model images as continuous functions and learn a mapping from spatial coordinates to latent representations~\cite{essakine2024we,molaei2023implicit,wolterink2022implicit}. INRs have shown advantages in modeling anatomical structures. In cardiac and brain MRI segmentation tasks with stable anatomy, prior work introduces subject-specific latent representations~\cite{stolt2023nisf,stolt2026nisf++} or meta-learning strategies~\cite{vyas2025fit}, enabling a shared implicit network to generalize to unseen subjects without explicit encoding. The success of these methods relies on structural consistency of anatomy. However, whether INRs can represent highly diverse pathological regions without well-defined structural patterns remains unclear. Unlike anatomical structures, WSIs lack a unified structural template and exhibit substantial structural variations in tissue, making INR formulations relying solely on point-wise coordinate inputs without modeling local spatial relationships insufficient for capturing complex tissue patterns. Although some INR methods enhance coordinate representations by enriched encodings~\cite{tancik2020fourier,mildenhall2021nerf}, they adopt global, fixed schemes that overlook local spatial structure, limiting their ability to model structural heterogeneity in WSIs. Multi-resolution hash grid coordinate encodings introduce learnable local feature tables that enable spatially adaptive allocation of representational capacity, making them well suited for capturing local structural variations in WSIs~\cite{muller2022instant,lee2024convolutional}.

Based on the above analysis, \textbf{we propose WSI-INR, a patch-free WSI learning framework based on INR}. Instead of treating a WSI as a set of discrete patches, WSI-INR models each WSI as a continuous function that maps spatial coordinates to image features and segmentation probabilities (Fig.~\ref{figure1}-(b)). Specifically, (1) WSI-INR directly predicts segmentation from spatial coordinates, preserving spatial information across the whole slide. (2) To address cross-resolution inconsistency, we introduce a multi-resolution hash grid encoding. Since the hash grid encoding performs multi-density spatial sampling of the same tissue, naturally matching the multi-resolution structure of the WSI pyramid, it enables consistent feature representations across different resolutions, thereby achieving more stable cross-scale segmentation (See Fig.~\ref{figure1}-(c)). (3) WSI-INR employs a shared decoder trained across multiple WSIs, enabling the model to learn shared morphological priors and generalize to unseen WSIs.

\textbf{Our contributions can be summarized as follows:} (1) We propose WSI-INR, which models a WSI as a continuous implicit representation. (2) WSI-INR models different resolutions as varying sampling densities of a single continuous function, enabling robust segmentation across resolutions. (3) Experimental results show that WSI-INR achieves superior robustness in cross-resolution prediction compared to classical patch-based methods. (4) We show that INRs can effectively segment pathological regions, not only structured anatomical regions.

\section{Method}
\begin{figure*}[tb]
  \centering
  \includegraphics[page=2, width=1\textwidth, clip, trim=0.8cm 10cm 2.3cm 0.5cm ]{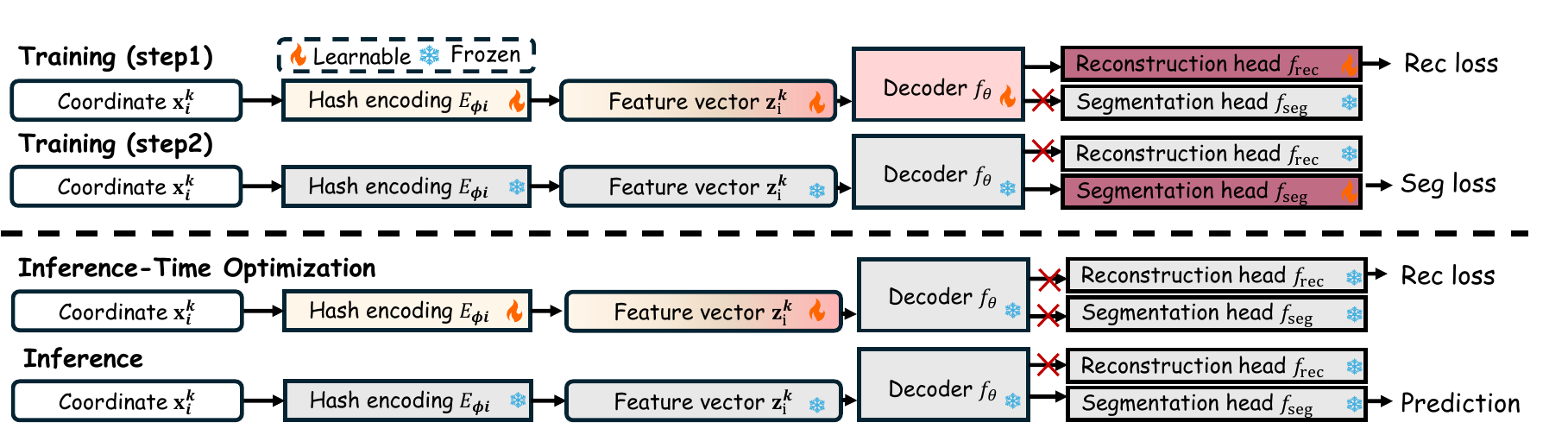}
  \caption{Workflow of our WSI-INR. Training: We first optimize the decoder, reconstruction head and hash encoding by minimizing the reconstruction loss. These modules are frozen, and only the segmentation head is trained. Inference: For each unseen WSI, network parameters remain fixed, while the hash encoding is optimized via the reconstruction loss. The final segmentation is obtained from the optimized representation.}
  \label{figure2}
\end{figure*}

\textbf{WSI-INR Overview.} In WSI-INR, for the $j$-th sampled local coordinate region on the $i$-th WSI, 
denoted as $\mathcal{N}(\mathbf{x}_i^{\,j}) \subset [0,1]^2$, 
the dense coordinates within a neighborhood are encoded to 
hash feature $\left\{
\mathbf{z}_i^{\,k} = E_{\phi_i}(\mathbf{x}_i^{\,k})
\;\middle|\; \mathbf{x}_i^{\,k} \in \mathcal{N}(\mathbf{x}_i^{\,j})
\right\}$,  $\mathbf{z}_i^{\,k}\in\mathbb{R}^{d_z}$. The feature vector $\mathbf{z}_i^{\,k}$ is fed into a shared decoder $f_{\theta}$ which is formulated as $ f_\theta : \mathbb{R}^{d_z} \rightarrow \mathbb{R}^{d_h}, \mathbf{h}_i^{\,j} = f_\theta(\mathbf{z}_i^{\,k}).$
The $\mathbf{h}_i^{\,k}\in\mathbb{R}^{d_h}$ denotes the intermediate representation and $d_z$ and $d_h$ are the feature dimensions. Finally, the vectors $\mathbf{h}i^{,k}$ are fed into the reconstruction head $f{\mathrm{rec}}$ and the segmentation head $f_{\mathrm{seg}}$.

% During inference-time optimization, update only the encoder $E_{\phi_i}$ and latent representations $\mathbf{z}_i(\cdot)$ and $\mathbf{h}_i$ using the reconstruction loss. The adapted model is then directly used to segment lesion areas on the WSI.

\noindent\textbf{Model Architecture Details.} 
\textit{Multi-resolution Hash Encoding.} We adopt multi-resolution hash encoding~\cite{muller2022instant} to encode continuous spatial coordinates, which has been shown to effectively enhance the capacity of implicit models in representing details and multi-resolution features~\cite{muller2022instant,lee2024convolutional}. In addition, WSIs naturally contain multi-scale structural patterns,
the hash grid representation provides an excellent match to the intrinsic pyramid structure of WSIs, enabling the model to simultaneously capture fine-grained textures and large-scale tissue structure (See Fig.~\ref{figure1}-(c)). We adopt a
WSI-specific hash encoding $E_{\phi_i}$ for each slide, since the spatial arrangement of tissue differs across WSIs. The encoding consists of $L$ grid levels, where the resolution of the $l$-th level is $R_l = R_0 \cdot s^{\,l},$ with $R_0$ the base resolution and $s$ the per-level scale factor. At each level $l$, the coordinate
$\mathbf{x}^j_i$ is scaled to resolution $R_l$ and its $K{=}2^d$
surrounding grid vertices are identified. When the vertex count exceeds the hash table capacity $T$, vertices are mapped to entries via a spatial hash function $h:\mathbb{Z}^2 \to \{0,\dots,T{-}1\}$, introducing collisions whose gradients average out during training. This effectively allocates more representational capacity to structurally complex regions while compressing homogeneous backgrounds, which is particularly suited to WSIs where informative structures are spatially sparse. The encoded feature at level $l$ is:
\begin{equation}
\mathbf{e}_{i,l}^{\,j}=\sum_{k=1}^{K}w^{(l)}_k(\mathbf{x}^j_i)\;\mathbf{T}^{(l)}\!\big[h(\mathbf{v}_k)\big],
\end{equation}
where $\mathbf{T}^{(l)}\!\in\!\mathbb{R}^{T \times F}$ is the learnable hash table, $w^{(l)}_k$ the interpolation weight and $F$ denotes the number of feature dimensions per level, and
$\mathbf{v}_k$ is the $k$-th neighboring vertex. The final encoding
concatenates all levels:
\begin{equation}
  \mathbf{z}^j_i = \bigoplus_{l=1}^{L}  \mathbf{e}_{i,l}^{\,j},
  \quad \mathbf{z}^j_i \in \mathbb{R}^{d_z},\;d_z = L \times F.
\end{equation}
\textit{Dual-branch Decoder.} Given the encoded hash feature vectors $\mathbf{z}_{i}^{\,j}$, we adopt a shared decoder~\cite{lee2024convolutional,sitzmann2020implicit} that captures both local spatial structures and global functionality relationships to map coordinate-based features into a unified implicit feature space . The decoder consists of two complementary branches: (1) a CNN branch that processes the encoded coordinate features to explicitly model local spatial continuity and neighborhood-level patterns, thereby capturing fine-grained tissue structures; and (2) an MLP branch that operates point-wise while capturing global spatial relationships through its implicit representation. The outputs of the two branches are fused at the feature level to form a unified implicit representation. 
This fused feature integrates local spatial constraints 
with global functional modeling capability, resulting in the final representation as $\mathbf{h}_i^{\,k} 
= \mathrm{Decoder}(\mathbf{z}_i^{\,k}).$

\textit{Reconstruction and Segmentation Heads.} Both the reconstruction and segmentation heads adopt a dilated residual architecture composed of residual blocks with progressively increasing dilation rates to enlarge the receptive field. On the fused implicit representation vector $\mathbf{h}_i^{\,j}$, a reconstruction head predicts image intensity from coordinates to regularize appearance modeling, while a segmentation head outputs lesion and background probabilities as follows:
\begin{equation}
\hat{\mathbf{s}}_i^{\,k} = f_{\mathrm{seg}}(\mathbf{h}_i^{\,k}) \in [0,1]^2,
\qquad
\hat{\mathbf{r}}_i^{\,k} = f_{\mathrm{rec}}(\mathbf{h}_i^{\,k}) \in [0,1]^3.
\end{equation}
\noindent\textbf{Training.}
As shown in Fig.~\ref{figure2}, we propose a two-step training strategy. This is because when reconstruction and segmentation are jointly optimized from the beginning, the segmentation objective quickly dominates, leading to shortcut learning that amplifies inter-class differences without capturing true tissue structures. Although the segmentation loss decreases rapidly, the model fails to learn meaningful lesion–normal representations. Therefore, we first learn a stable implicit image representation and then introduce segmentation supervision. This allows the segmentation head to learn discriminative features within a semantically structured feature space for accurate lesion segmentation. In step 1, the encoder parameters $\phi$, decoder parameters $\theta$, and reconstruction head are optimized while the segmentation head is frozen. The model is supervised by the reconstruction loss (Mean Squared Error) as follows:
\begin{equation}
\mathcal{L}_{\text{stage1}}
=
\mathcal{L}_{\text{rec}}\!\left(
f_{\text{rec}}(\mathbf{x}_i^{\,k}),
\mathbf{I}(\mathbf{x}_i^{\,k})
\right).
\label{equa1}
\end{equation}
where $\mathbf{I}(\mathbf{x}_i^{\,k})$ denotes the image intensity. This stage establishes an implicit WSI representation to model continuous appearance and spatial structures from WSIs. In step 2, where all learned components are frozen and only the segmentation head is optimized by binary cross-entropy (BCE) and Dice~\cite{milletari2016v} loss:
\begin{equation}
\mathcal{L}_{\text{stage2}}
=
\mathcal{L}_{\text{BCE}}\!\left(
f_{\text{seg}}(\mathbf{x}_i^{\,k}),
\mathbf{S}(\mathbf{x}_i^{\,k})
\right)
+
\mathcal{L}_{\text{Dice}}\!\left(
f_{\text{seg}}(\mathbf{x}_i^{\,k}),
\mathbf{S}(\mathbf{x}_i^{\,k})
\right).
\end{equation}
where $\mathbf{S}(\mathbf{x}_i^{\,k})$ denotes segmentation labels. 

\noindent\textbf{Inference-time Optimization (ITO) and Inference.} During the ITO stage, an unseen WSI requires optimizing its corresponding hash encoding. We freeze all global network and optimize only the hash encoding specific to the target WSI, allowing the model to adapt to it using the reconstruction objective without any segmentation supervision, consistent with the ITO protocol. This process preserves the global network parameters while enabling the hash grid to rapidly adapt to the texture and structural characteristics of the new slide, while retaining the semantic priors learned by the shared decoder during training. The optimization objective is the same as in Eq.~\ref{equa1}. After completing ITO, we perform dense prediction over the entire WSI by querying spatial coordinates in the continuous domain at the desired resolution and forwarding them through the network to obtain segmentation results.

\begin{figure*}[tb]
  \centering
  \includegraphics[page=3,width=1\textwidth, clip, trim=0cm 2.5cm 0cm 2.3cm ]{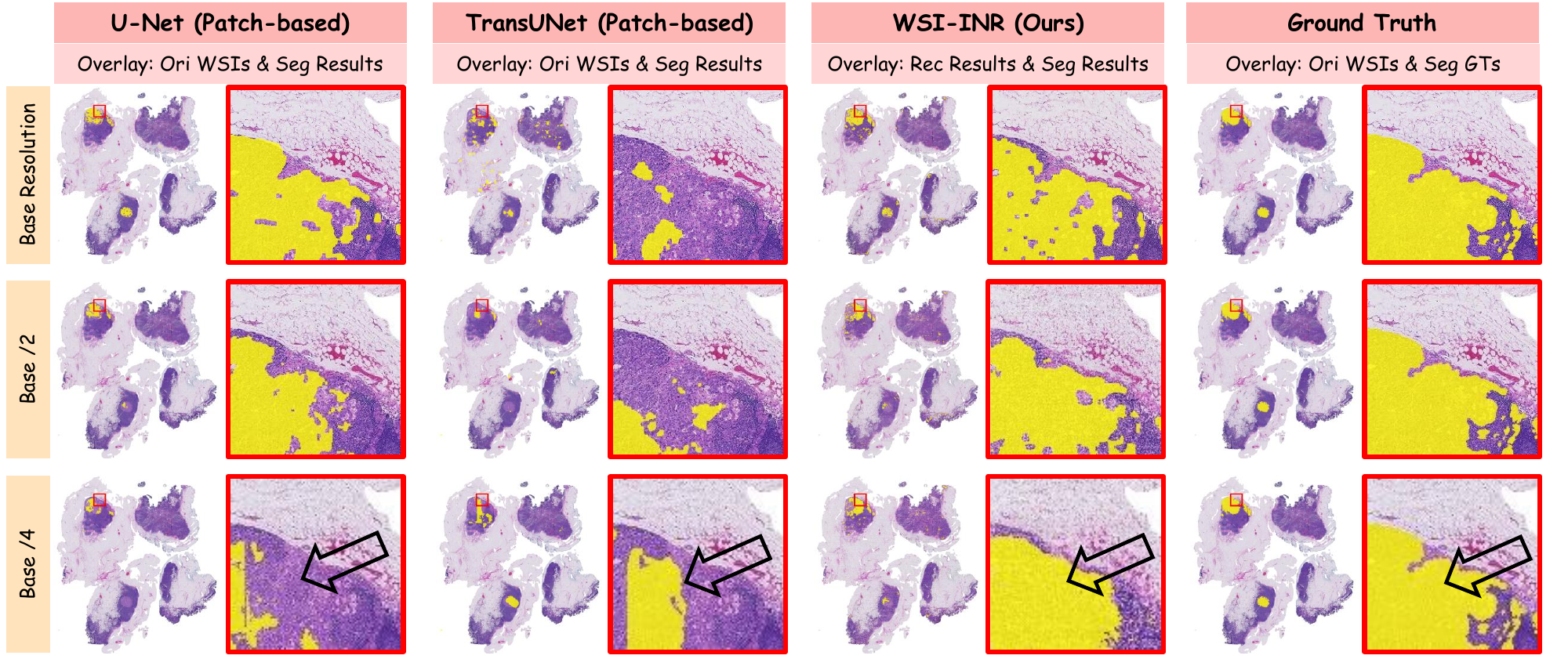}
  \caption{We compare our WSI-INR with U-Net  and TransUNet across multi resolutions, while all models are trained at a single base resolution. For WSI-INR, we adopt base-resolution optimization. U-Net performs well at training resolution, but U-Net and TransUNet degrade at lower resolutions, with fragmented predictions (black arrows).}
  \label{figure3}
\end{figure*}

\section{Experiments}
\label{sec:experiment}
\textbf{Dataset and Metrics.}We conduct experiments on an H\&E-stained WSI datasets for breast cancer lymph node metastasis detection, CAMELYON16~\cite{bejnordi2017diagnostic,litjens20181399}. Training is performed on the 39 WSIs containing macro-metastases in CAMELYON16. For evaluation, we use 22 WSIs with macro-metastases from the official CAMELYON16 test set. We follow~\cite{stolt2023nisf} to use Dice as the evaluation metric.

\noindent\textbf{Implementation Details.}
Our method is implemented using PyTorch~\cite{paszke2019pytorch} and the TinyCUDA-NN framework~\cite{muller2022instant}. 
Input coordinates are normalized to the continuous domain $[0,1]^2$ while preserving the aspect ratio of each WSI, and pixel intensities are normalized to the range $[0,1]$. The multi-resolution hash grid contains $L=21$ levels with a base resolution $R_0$=16 and a per-level scale factor $s$=1.5. Each level maintains a hash table of size $T = 2^{21}$ with a feature dimension of $F=2$. The model is trained for 200 epochs, with the first 100 epochs dedicated to reconstruction and the remaining epochs used for segmentation. 
We use the Adam optimizer~\cite{kingma2014adam} with a learning rate of $1\times10^{-5}$. The batch size is set to 1, and each batch samples dense spatial coordinates within a $1024\times1024$ window of WSIs. Our method is implemented on NVIDIA A100 GPUs, and the total training time is about 20 hours. For each WSI, we perform ITO with a maximum of 20 epochs at the base resolution. Resolution-specific optimization follows the same configuration at their respective resolutions. ITO adopts a dual-criterion dynamic early-stopping strategy with Exponential Moving Average (EMA). Optimization stops when MSE < 0.002; after a 5-epoch warm-up, it terminates if the current MSE exceeds the historical minimum by 30\%. EMA parameters are maintained throughout optimization, and EMA weights are used for final inference to ensure stable and smooth predictions. The inference model contains 42.3M parameters. The processed WSIs range from 5,376–9,984 × 5,152–13,856 pixels, with an average total ITO and inference time of 6.12 minutes per WSI.

\begin{table*}[t]
\centering
\caption{Segmentation performance on CAMELYON16 across resolutions evaluated by Dice score. Models are trained at the base resolution (Base) and evaluated at the same resolution as well as two other unseen resolutions in training (Base/2 and /4). Values in \textcolor{red}{red} denote the performance change relative to the results at the base resolution.}
\label{tab:main_results}

\resizebox{\textwidth}{!}{
\begin{tabular}{lccc}
\toprule
\textbf{Method} 
& \textbf{~Resolution: Base~} 
& \textbf{~Resolution: Base/2~} 
& \textbf{~Resolution: Base/4~} \\
\midrule

U-Net~\cite{ronneberger2015u} 
& 0.4858  % 这里填 Center0和1的平均值
& 0.2418 \textcolor{red}{(-50.23\%)} % 填合并后的值和新算的下降率
& 0.2221 \textcolor{red}{(-54.28\%)} \\
TransUNet~\cite{chen2024transunet} 
& 0.1534
& 0.1146 \textcolor{red}{(-25.29\%)} 
& 0.0979 \textcolor{red}{(-36.18\%)} \\
\midrule
\textbf{Ours (Base-resolution opt)} 
& \textbf{0.2417} 
& \textbf{0.1664} \textbf{\textcolor{red}{(-31.15\%)} }
& \textbf{0.1683} \textbf{\textcolor{red}{(-30.37\%)}} \\
\textbf{Ours (Resolution-specific opt)} 
& \textbf{0.2417} 
& \textbf{0.2333} \textbf{\textcolor{red}{(-3.48\%)} }
& \textbf{0.3048} \textbf{\textcolor{red}{(+26.11\%)}} \\
\bottomrule
\end{tabular}
}
\end{table*}

\section{Experimental Results and Discussion}
\label{sec:result}
\noindent\textbf{WSI Segmentation Cross-resolution Evaluation.} We evaluate WSI-INR on the CAMELYON16 dataset under cross-resolution testing settings, comparing it with classical patch-based methods (U-Net~\cite{ronneberger2015u} and TransUNet~\cite{chen2024transunet}). All methods are trained at a high base resolution and directly evaluated at the base scale and lower scales (Base/2 and Base/4). We also set a resolution-specific optimization, where ITO is performed independently at each target resolution. As shown in Table~\ref{tab:main_results}, patch-based methods exhibit clear performance degradation across resolutions, indicating sensitivity to resolution changes. In contrast, WSI-INR maintains stable performance, benefiting from its continuous representation that accommodates spatial sampling variations. Under resolution-specific optimization, WSI-INR outperforms TransUNet and achieves performance close to U-Net at Base/2 and Base/4. Despite multi-scale designs, U-Net and TransUNet remain sensitive to resolution shifts and produce spatially discontinuous predictions (Fig.~\ref{figure3}), whereas WSI-INR preserves structural continuity.

\begin{table*}[t]
\centering
\caption{Ablation study for coordinate encoding on CAMELYON16. Models are trained at base resolution and evaluated at the same resolution and other resolutions.}
\label{tab:ablation}

% 增加列间距，让表格自然撑开
%\setlength{\tabcolsep}{15pt}

\resizebox{\textwidth}{!}{
\begin{tabular}{lccc} 
\toprule
\textbf{Method} 
& \textbf{~~Resolution: Base~~} 
& \textbf{~~Resolution: Base/2~~} 
& \textbf{~~Resolution: Base/4~~} \\
\midrule

w/o encoding 
& 0.0000 
& 0.0000 
& 0.0000 \\

w/ NeRF positional encoding~\cite{mildenhall2021nerf} 
& 0.0901 % (0.044+0.157)/2
& 0.0901 
& 0.0901 \\

\midrule

\textbf{w/ hash grid encoding} 
& \textbf{0.2417} 
& \textbf{0.1664} 
& \textbf{0.1683} \\

\bottomrule
\end{tabular}
}
\end{table*}

\begin{figure*}[tb]
  \centering
  \includegraphics[page=4,width=1\textwidth, clip, trim=0cm 8cm 0cm 0cm ]{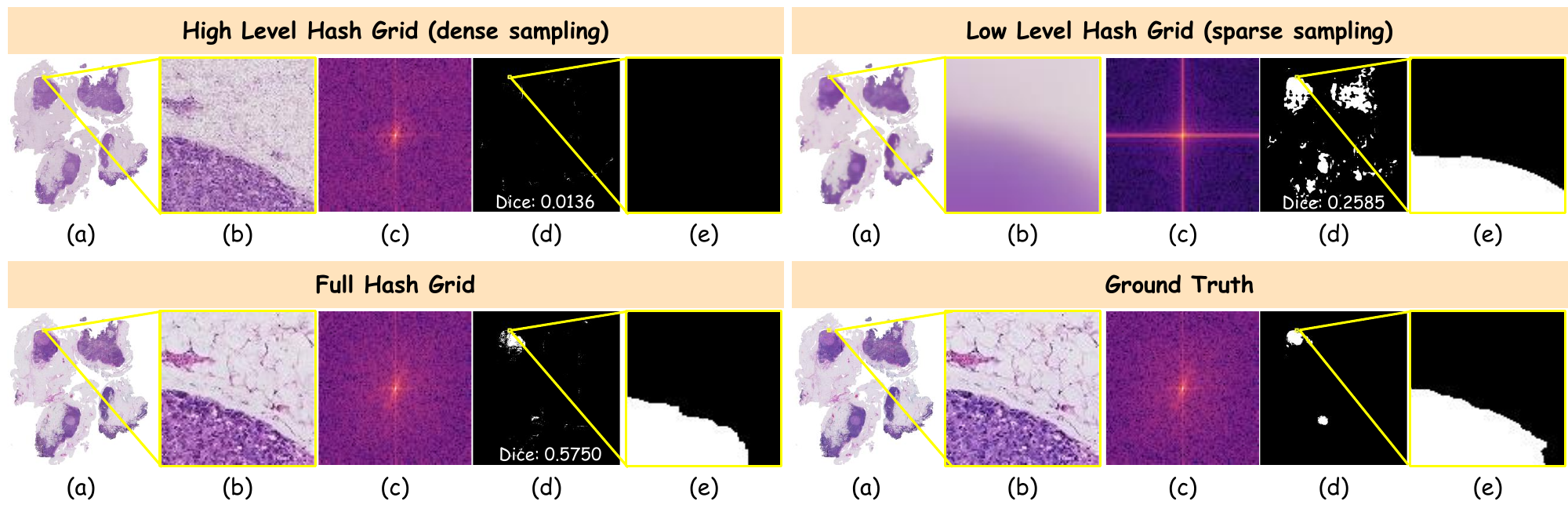}
  \caption{We visualize the reconstruction and segmentation results under three settings: high-level only (dense sampling), low-level only, and the full hash grid. Columns: (a) global reconstruction, (b) zoomed-in local patch, (c) frequency spectrum (FFT) of local patch, (d) global segmentation prediction, and (e) local segmentation mask.}
  \label{figure4}
\end{figure*}

\noindent\textbf{Analysis of Multi-Resolution Hash Encoding.} We analyze the effect of encoding by comparing models without encoding, with NeRF positional encoding~\cite{mildenhall2021nerf}, and with hash grid encoding. The experimental results in Table~\ref{tab:ablation} show that models without any encoding, or those using fixed and scale-coupled positional encodings like NeRF positional encoding, struggle to handle the high heterogeneity and complex structural patterns of WSIs, leading to poor reconstruction and lesion segmentation. As shown in Fig.~\ref{figure4}, during ITO we decouple the high- and low-level hash components to analyze their roles. High-level hash (dense sampling) captures high-frequency details but lacks spatial coherence, leading to segmentation failure. Low-level only preserves global structures but misses fine details such as cells. The full hash grid integrates both, achieving high-fidelity reconstruction and strong segmentation performance. These results indicate that segmentation performance depends on reconstruction quality, as the segmentation head relies on the learned implicit representation.

\section{Conclusion}
\label{sec:conclusion}
In this work, we propose WSI-INR, a patch-free framework for WSI lesion segmentation. By modeling the WSI as a continuous implicit function, WSI-INR directly predicts segmentation from spatial coordinates while preserving global spatial continuity. Multi-resolution encoding treats different resolutions as varying sampling densities of a shared latent function, enabling robust cross-resolution performance. Significantly, we extend the segmentation capability of INRs from structurally consistent anatomical structures to highly heterogeneous pathological lesions. Although limitations remain in modeling micro-scale lesion regions and in generalizing across multi-center datasets, we believe that WSI-INR provides a scalable framework for continuous representation and multi-scale modeling in computational pathology. Future work will further expand WSI-INR to additional WSI analysis tasks and improve cross-institutional generalization.

\section*{Acknowledgements}
This work was funded by the JST Moonshot R\&D Grant Numbers JPMJMS2033, Yunheng Wu were supported by the Nagoya University CIBoG WISE program from MEXT.

\bibliographystyle{splncs04}
\bibliography{MICCAI2026-main_conference_paper_template.bib}

\end{document}